\documentclass{ifacconf}
\usepackage{url}
\usepackage{graphicx}      
\usepackage{natbib}        
\usepackage{amsmath}
\usepackage{amssymb}
\usepackage{graphicx}
\usepackage{bbm}
\usepackage{svg}
\usepackage{comment}
\usepackage{subcaption} 
\usepackage[justification=centering]{caption}
\begin{document}
\begin{frontmatter}

\title{Learning Autonomy: Off-Road Navigation Enhanced by Human Input} 


\author[First]{Akhil Nagariya,Dimitar Filev,Srikanth Saripalli} 
\author[Second]{Srikanth Saripalli, Gaurav Pandey}

\address[First]{J. Mike Walker '66 Department of Mechanical Engineering, Texas A\&M University.}
\address[Second]{Department of Engineering technology and Industrial Distribution, Texas A\&M University}

\begin{abstract}                
 In the area of autonomous driving, navigating off-road terrains presents a unique set of challenges, from unpredictable 
    surfaces like grass and dirt to unexpected obstacles such as bushes and puddles. In this work, we present a novel learning-based local planner that addresses these challenges by directly capturing human driving nuances from real-world demonstrations using only a monocular camera. The key features of our planner are its ability to navigate in challenging off-road
    environments with various terrain types and its fast learning capabilities. By utilizing minimal human demonstration data (5-10 mins), 
    it quickly learns to navigate in a wide array of off-road conditions.
    The local planner significantly reduces the real world data required to learn human driving preferences. This allows the planner to apply learned behaviors to real-world scenarios without 
    the need for manual fine-tuning, demonstrating quick adjustment and adaptability in off-road autonomous 
    driving technology.
\end{abstract}

\begin{keyword}
Offroad Robot Navigation, Task and Motion Planning, Path Following
\end{keyword}

\end{frontmatter}

\section{INTRODUCTION}

   Off-road planning and navigation present unique challenges due to the unpredictable nature of various terrains and their geometric 
   characteristics. Successfully navigating these environments requires leveraging both visual and geometric features 
   effectively. Modeling tire-terrain interactions and vehicle dynamics across diverse off-road conditions is a complex task. 
   Even with accurate models, tuning the planning algorithm to navigate safely across different terrains demands extensive 
   time and expertise. In our research, we introduce a demonstration-based local planning algorithm that bypasses the need for 
   directly modeling these intricate dynamic interactions. Instead, it learns navigation preferences from human driving data, 
   demonstrating the ability to adapt these learned behaviors from simulations to real vehicles with minimal manual adjustments.

   Our approach uses utility \cite{DAI2022103916} functions to directly extract key features from segmented images and learns human driving behaviour using demonstration data. This approach diverges from traditional methods, which typically require either extensive labeled data for 
   end-to-end learning or precise sensor calibration and global mapping in classical robotics approaches. 
   By focusing on extracting key features directly in the trajectory space, our method simplifies the process, 
   avoiding the complexity of global map generation. This allows our algorithm to effectively learn driving patterns from as 
   little as 5-10 minutes of driving data. Our approach reduces the need for large datasets and detailed calibration, facilitating 
   a more straightforward and efficient learning algorithm for navigating complex environments.
    The main contributions of this work are outlined below:
    \begin{itemize}
    \item Fast learning capabilities: The planner can learn complex navigation behaviors from as little as 5-10 minutes of human demonstration data, significantly reducing the data requirements compared to traditional approaches.
    \item Adaptability across diverse terrains: Our algorithm demonstrates the ability to navigate effectively in a wide array of off-road conditions, including mud, rock, water, and non-traversable areas 
    \item Human-like decision making: By learning from human demonstrations, our planner exhibits more intuitive and human-like navigation choices when faced with complex terrain configurations
    \item Reduced need for manual tuning: Our approach bypasses the need for extensive manual tuning of cost functions or precise modeling of vehicle-terrain interactions, making it more accessible for deployment in diverse environments
    \end{itemize}

\section{Related Work} 
Classical work in off-road navigation 
\cite{kellyReliableRoadAutonomous2006,kimTraversabilityClassificationUGV2007} has focused on creating the costmap
of the environment from sensors' data to represent navigation cost associated
with the various types of visual and geometric features of the environment. 
Earlier approaches \cite{kellyReliableRoadAutonomous2006,thrunWinningDARPAGrand2006}
relied on feature engineering while later approaches
\cite{maturanaRealTimeSemanticMapping2018,valadaDeepMultispectralSemantic2017} relied on deep learning based 
semantic segmentation to represent the visual and geometric features of the surrounding terrain. These later approaches 
train the semantic segmentation pipeline from scratch and use hand designed cost functions for the planner. 
Although, these costmaps provide rich information for the downstream planning tasks, 
tuning them to capture the complex dynamic interactions while navigating on various terrain types is extremely challenging and requires significant domain expertise.

Recent advances in Deep learning 
have inspired researchers in robotics community to develop end-to-end learning algorithms\cite{bajracharyaAutonomousRoadNavigation2009,kahnBADGRAutonomousSelfSupervised2020} that 
directly learn the mapping from sensor information to control commands thereby bypassing 
the need for manual costmap creation and tuning. Despite the promises of these end-to-end approaches, they still require a large amount of data and show poor generalisation to different settings (domain adaptation). Moreover, the black box nature of these approaches make them very hard to debug and deploy on real systems.

To deal with these challenges, recent research has focused on learning-based algorithms that combine the strengths of classical and more recent end to end approaches. These efforts aim towards an acceptable trade off between domain expertise, explain-ability and training data requirements. 

Recent works like \cite{triestLearningRiskAwareCostmaps2023, caiProbabilisticTraversabilityModel2023a}
leverage human driving data to either directly learn the costmap of the environment or learn the traversability of various terrain types. 
These costmaps are crucial to solve the optimal control problem and generate appropriate vehicle controls. While these methods effectively model the tire-terrain interactions of the vehicle, they often overlook the higher level reasoning required to navigate in challenging off-road environments. Additionally, these methods primarily depend on real-world data to learn costmaps and traversability. In contrast, our research demonstrates the potential of simulation platforms to learn a local planning algorithm. 

\section{Problem Statement}
Given a predefined, ordered set of waypoints in an off-road environment, our objective is to develop a local planning algorithm that enables a vehicle to follow these waypoints as accurately as possible. In doing so, the algorithm should incorporate human preferences in selecting the type of terrain the vehicle traverses. In the following section we provide some definitions that are necessary to 
formalize what we mean by human preferences, but before we do that we discuss some of the assumptions made in this work:
\begin{itemize}
\item Off-road environment: planar geometry, consists of 4 terrain types: Non traversable (trees, big rocks etc), water, rock and mud.
\item Vehicle type: Differential drive kinematics, accepts linear and angular velocities as commands.
\item We consider the discrete case in this work with a step size of $\Delta t$
\end{itemize}

\subsection{\textbf{Definitions}}
\begin{itemize}
    \item \textbf{Pose}: A pose represents vehicle position and orientation in a
local frame and is given by a 3-tuple $(x, y, \theta)$, where $(x, y)$
is the position of the vehicle in a 2-D local frame and $\theta$ is
the orientation.
\item \textbf{Reference path}: Reference path ($P) $ is an ordered set of poses that 
are collected by manually driving the vehicle in off road environment. 
\item \textbf{Control Commands}: The control command
for the vehicle is represented by a 2-tuple $(v,w)$, where $v$ is the linear velocity and $w$
is the angular velocity commands.
    \item \textbf{Trajectory}:  We define a Trajectory $T$ as a
finite set of ordered poses: $T = [\{x_t,y_t,\theta_t\}_{t = 1}^{t = n}]$. In this work, we consider trajectories of fixed length. The poses in the trajectory are assumed to be generated by the vehicle (or its kinematic model) at a fixed time step 
$\Delta t$, while executing a constant control command over the entire horizon of 
$n$ steps. This results in a one-to-one mapping between the control commands and the trajectory at \textit{every} instant.
    \item \textbf{Human preference}: At any given moment, human preference is defined by the control commands chosen while following a predefined set of waypoints. Given the one-to-one correspondence between control commands and trajectories, human preference can alternatively be described by the trajectory selected at that instant.
    \item \textbf{Preference set $S$:}
The preference set $S$ is a fixed set that represents all the different preferences (Trajectories) available
for the human operator at any instant. To construct $S$ we first discretize the control commands—linear and angular velocities. Let  $A$, represent this fixed discrete set of linear and angular velocities:
\begin{align}
   A = [\{v^i, w^i\}_{i = 1}^{i= m}] \label{eq:control}
\end{align}

Assuming the vehicle starts at 
$[0,0,0]$, we then apply forward simulation using the differential drive kinematic model to generate $m$ distinct trajectories over a pre defined time horizon $n$.
\begin{align*}
    T^i &= [\{x^i_t, y^i_t, \theta^i_t\}_{t=1}^{t=n}] \\
    x^i_t &= x^i_{t-1} + v^i \Delta t cos(\theta^i_{t-1}) \\
    y^i_t &= x^i_{t-1} + v^i \Delta t sin(\theta^i_{t-1}) \\
    \theta^i_t& =  \theta^i_{t-1} +w^i \Delta t \\
    \{x^i_1, y^i_1, \theta^i_1\} &= \{0,0,0\} \quad  \forall i \in \{1,..,m\}
\end{align*}
The set $S = \{T^i\}_{i=1}^{i=m}$ is defined as the preference set. We also refer to $A$ as the control set. Both the preference set
and the control set are indexed sets where an index is used to identify the control command/trajectory uniquely. In this work we are only interested in 
learning the lateral control for the vehicle and fix $v^i$s to 1m/s, ($\forall i \in \{1,..,m\})$.  
\end{itemize}
\section{METHODOLOGY}
To characterize the various features of the local environment and associating them to human preferences,
we introduce "utility feature" ($U(S, P) \in \mathbb{R}^{m\times5}$), defined for 
a preference set (S) and a reference path (P): 
\begin{align}
U(S,P)
&=
\begin{bmatrix}\label{eq:uf}
  u^1(T^1) & u^2(T^1) &\ldots &u^5(T^1,S, P)\\
  u^1(T^2) & u^2(T^3)&\ldots &u^5(T^2,S, P)\\
  \vdots & \vdots & \vdots   & \vdots \\
  u^1(T^m) & u^2(T^m) &\ldots &u^5(T^m,S, P)
\end{bmatrix} 
\end{align}
\begin{center}
$T^i \in S, \hspace{4pt} \forall i \in 1,..m$
\end{center}
Here, $u^1(T),..,u^4(T)$ are utility functions associated with four different terrain features
of the environment and depend only on a single trajectory. The utility function $u^5(T,S,P)$ represents the distance-based utility, which depends on a trajectory $T$, the preference set $S$ and the reference path $P$. Each row $i$ in Eq.~\ref{eq:uf} corresponds to all utility functions for the trajectory $T^i \in S$.
In the next section, we discuss these utility functions in detail.

\subsection{Utility Functions} \label{utility}
In this work, we consider five utility functions, which are categorized into terrain utility functions and distance utility functions. The first four, $u^1(T)$ through $u^4(T)$, are terrain utility functions that correspond to different terrain types in the environment: non-traversable, water, rock, and mud, respectively. The fifth utility function, 
$u^5(T, S, P)$,is a distance-based utility function that measures how close a trajectory $T$ is to a reference path $P$, relative to the other trajectories in the preference set $S$. To calculate the utility functions corresponding to the
4 terrain types, we project the trajectory T on to the camera plane and then
use the following equation.
\begin{align}
    u^k(T) = u^k(g(T)) = \frac{1}{n} \sum_{j=1}^{j=n} \mathbf{1}^k{(I(j))} \label{eq:2}
\end{align}
Here $k \in {1,2,3,4}$ is the identifier index for the four different terrain types
(1-non traversable, 2-water, 3-rock, 4-mud). The projection of $T$ on to the 
camera plane is given by $g(T)$. $I(j) \in \{1,2,3,4\}$, is the pixel label(in the segmented image) that intersects 
with the $j^{th}$ point in the trajectory and $\mathbf{1}^k$ is the indicator
function:
\begin{align}
    \mathbf{1}^k(x) = 
    \begin{cases}
      1 & \text{if $x=k$} \\
      0 & \text{otherwise}
    \end{cases}
\end{align}
The distance utility function is defined with respect to a preference set and a reference path. 
Given a preference set $S = \{T^i\}_{i = 1}^{i = m}$, for each trajectory  $T^i \in S$, let  $d(T^i, P)$ denote the distance between the final point of the trajectory and the nearest waypoint on the reference path
$P$. Using this, we define the distance utility function $(u^5(T^i, S, P))$:
\begin{align}
    u^5(T^i, S, P) =1 - \frac{d(T^i, P) - d_{\min}}{d_{\max} - d_{\min}} \label{eq:4}
\end{align}
where:
\begin{align*}
   d_{\min} &= \min\limits_{T^j \in S} d(T^j, P) \\
   d_{\max} &= \max\limits_{T^j \in S} d(T^j, P)
\end{align*}
Thus the distance utility function represents the relative closeness of a trajectory from the path $P$ within the preference set $S$.

\subsection{Learning human preferences using demonstrations}
To learn human preferences we frame the problem as a muticriteria decision-making problem where, at each time step, the planner evaluates a set of utilities representing the degree of satisfying multiple criteria (specific terrain preference and closeness to the reference path) and selects a trajectory from a fixed preference set S of trajectories. Since we are dealing with multiple utilities associated with a specific trajectory, we are interested in an overall decision function that aggregates the utilities $(u^1(T, S,P), .., u^5(T, S, P))$ and indicates the planner's degree of preference for that possible trajectory. To avoid the cumbersome process of defining the structure and parameters of the decision function \cite{DAI2022103916} we transform the muticriteria decision-making problem to a classification problem.
The labels for this classification task are derived from human demonstrations, where at each time step, the human operator selects a trajectory from the same fixed preference set 
$S$ while following a reference path. During this demonstration, the human operator follows the reference path as closely as possible, while simultaneously avoiding or preferring different terrain types based on the real-time camera feed displaying the environment ahead of the vehicle. In the following subsections we define various terms required to formalize the supervised learning problem.

\subsubsection{Trajectory Labels}
During the data collection the human operator chooses a control command from the set $A$ (Eq.~\ref{eq:control}) at every instant. The index of this control command uniquely identifies the trajectory/control preference and provides us with the ground-truth label for that instant. The index labels are then converted to one-hot-encoded vectors for supervised learning training. For an instant $t$, we use $L_t$ to represent the corresponding ground truth one-hot-encoded vector.

\subsubsection{Prediction}
Given the utility feature $U_t(S, P)$ at each instant $t$, the classifier outputs
the probabilities $F_t$ of all the trajectories in the reference set. We use Eq. \ref{eq:2} to calculate the first 4 terrain utility functions. 
To calculate the last utility function we first transform all the waypoints to the local vehicle frame and then use Eq. \ref{eq:4}. We finally use Eq. \ref{eq:uf} to calculate the utility feature $U_t(S, P)$. Please note that we only need to compute the preference set $S$ once since the utility feature is calculated in the local vehicle frame at each time instant $t$.

We implement the classifier using a neural network, denoted as 
$C$, as shown in Fig. \ref{fig:nn}. $C$ maps the utility feature $(U_t(S,p)$ to the vector of predicted probabilities ($F_t$) of all the trajectories in the preference set $(S)$. The input layer in $C$ compresses the utility features into an 
$m\times1$ feature vector, which is then transposed and passed through a block of three fully connected hidden layers, followed by another transpose and a softmax operation: 
\begin{figure}
\captionsetup{font=footnotesize}
\centering
    \includesvg[width=0.5\textwidth]{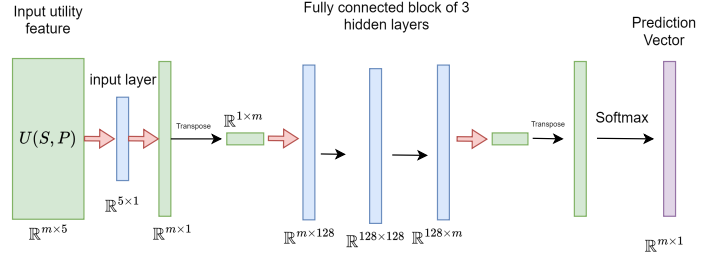}
    \caption[]{Architecture of the classifier. Layers and data are represented by blue and green rectangles respectively. Here $m=21$, Since we have 21 different trajectories in the 
    preference set $S$ (Section \ref{pref})}
    \label{fig:nn}
\end{figure}
\begin{align}
    F_t = C(U_t(S,P)) \label{eq:6}
\end{align}

\subsubsection{Data collection} \label{feature}
Given a reference set of waypoints the human operator is asked to drive the vehicle in the off road environment.
At every instant $t$ we record the ground truth labels $L_t$ and calculate the prediction vector $F_t$ using Eq. \ref{eq:6}.
We aggregate this data to  construct the training dataset $\mathbf{D}$ :  
\begin{align}
\mathbf{D} = [\{F_t, L_t\}_{t=1}^{t=N}] \label{eq:7} 
\end{align}
where $N$ is total points in $\mathbf{D}$. 

\subsubsection{Learning objective}
Given the dataset $\mathbf{D}$ (Eq. \ref{eq:7}) we use the cross entropy loss to minimize 
the error between the predicted probabilities and the ground truth labels:
\begin{align}
\min  \sum_{t=1}^{N} -L_t^T \log(F_t) \label{eq:8}
\end{align}

\section{Training and Evaluation} \label{pref}
\subsection{Simulation Setup}
We evaluated our algorithm using the AirSim \cite{airsim2017fsr} simulation environment. We developed a Warthog unmanned ground vehicle (UGV) model and created an off-road environment with four distinct terrain types: mud, rock, water, and non-traversable. We attach a front facing
camera and a GPS sensor on the warthog to get the segmentation images and the vehicle pose respectively. The camera is situated at a height of 1m from the ground and is tilted by 30 degrees downward to get a better view of the terrain in front.
Here are some of the details of the evaluation setup:
\begin{itemize}
\item The simulation environment provides semantic segmentation images of size $(640\times480)$
at 10Hz. 
\item The internal controller of the warthog expects linear and angular velocities as control commands at 30Hz.
\item A joystick is used to drive the warthog manually in the environment which 
provides linear velocities in range of (0, 1m/s)  and angular velocity in the range of (-1 rad/s to 1 rad/s). 
\item The reference path is collected by manually driving the vehicle around in the environment.
\item During the training run, we discretized the joystick’s angular velocity into 21 bins with a resolution of 0.1 rad/s while maintaining a constant linear velocity of 1m/s. This discretization resulted in 21 distinct trajectories within the preference set $S$
\end{itemize}
Given the inherent subjectivity of human driving behavior, we establish a terrain preference hierarchy to properly evaluate our algorithm:
\begin{align}
mud > rock > water > non \hspace{2mm} traversable \label{eq:9}
\end{align}
This qualitative ordering implies that a human driver would prefer to traverse muddy terrain over rocky terrain, choose rocky terrain over water, and avoid non-traversable terrain entirely. By defining this preference order, we can effectively assess our algorithm's performance in replicating human-like decision-making across different terrains. 

To evaluate our algorithm, we structured the training and testing phases to progressively assess its generalization capabilities. \textit{During training, we provided scenarios to the human driver that required making choices between terrain types that are consecutive in relation~\ref{eq:9}}. This means the driver only encountered decisions between adjacent terrain preferences in our established hierarchy. \textit{In contrast, during testing, the algorithm was presented with scenarios involving terrains without any such restrictions, allowing for choices between non-consecutive terrain types.}

Furthermore, we altered the shapes of the terrain patches during testing to configurations not present during training. This variation ensures that the algorithm is evaluated on its ability to handle unfamiliar terrain shapes and combinations, thereby testing its robustness and adaptability beyond the trained scenarios 
\begin{figure*}[htbp]
\captionsetup{justification=centering,font=footnotesize}
    \centering
     \begin{subfigure}[b]{0.49\textwidth} 
        \centering
        \includesvg[width=\textwidth]{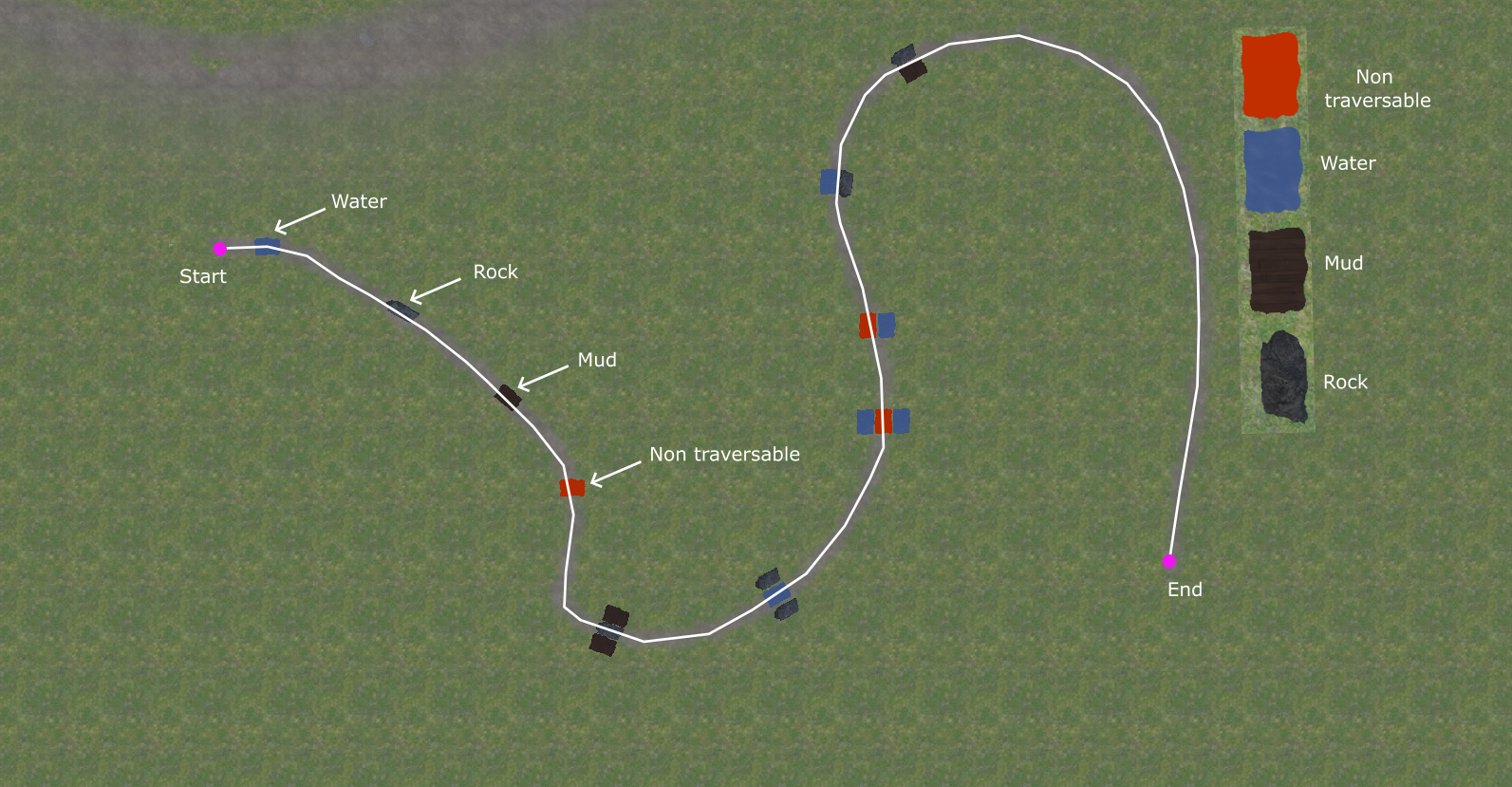}
        \caption{Top view of training environment. White curve is the reference path, different terrain types are also shown using arrows.}
        \label{fig:train_sim}
    \end{subfigure}
    \hfill
    \begin{subfigure}[b]{0.49\textwidth}
        \centering
        \includesvg[width=\textwidth]{figs/top_view_training_marked.svg}
        \caption{Red curve shows the human driven path providing 10 different examples for terrain preference}
        \label{fig:train_sim2}
    \end{subfigure}
    \caption{Top down view of the AirSim environment}
    \label{fig:main}
\end{figure*}
\begin{figure*}[htbp]
\captionsetup{font=footnotesize}
\centering
    \includesvg[width=0.6\linewidth]{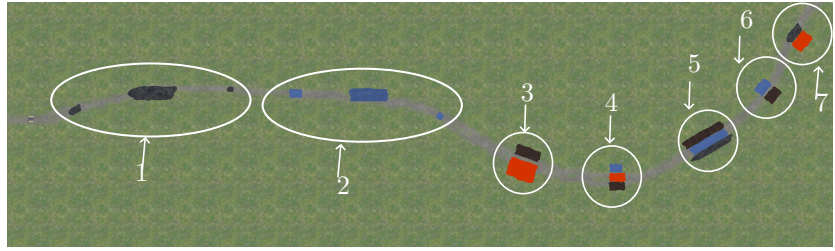}
    \caption[]{Shows the testing environment. Different scenarios are circled and
    are numbered from 1-7}
    \label{fig:test_sim}
\end{figure*}
\begin{figure*}[htbp]
\captionsetup{font=footnotesize}
\centering
    \includesvg[width=0.6\linewidth]{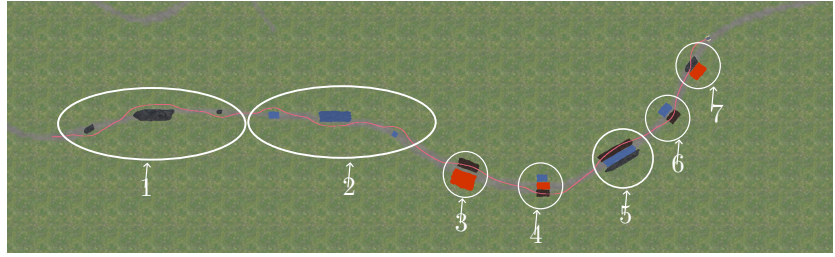}
    \caption[]{Shows the warthog trajectory (pink curve) in the testing scenarios. All the scenarios are shown by the white ellipses.}
    \label{fig:test_sim2}
\end{figure*}
\subsection{Training}

Fig. \ref{fig:train_sim} presents a top-down view of the training environment in AirSim. The reference path for the human operator, shown in white, corresponds to a trail within a grassy landscape that the operator is instructed to follow. This trail includes four terrain patches—water, rock, mud, and non-traversable areas—designed to present different scenarios to the operator. Although the grassy terrain is traversable, in this work we assume that we want to follow the waypoints as closely as possible so the warthog
cannot simply take very wide turns to avoid all the terrain types. Note that during the training (Fig. \ref{fig:train_sim}),
we only provide choices between terrain types that are consecutive in relation~\ref{eq:9}.

Fig. \ref{fig:train_sim2} shows the actual path navigated by the human operator during training. The scenarios where the operator deviates from the reference path are numbered from 1 to 10. The first four deviations occur in areas containing only a single terrain (no choice), which the operator simply avoids. In contrast, the subsequent six deviations involve scenarios with two or more terrain types, requiring the operator to make choices between them. The various terrains involved in scenarios numbered 1 through 10, along with the corresponding human preferences, are presented in Table \ref{tab:sample}

\begin{figure}
\captionsetup{font=footnotesize}
    \includegraphics[width=0.9\linewidth]{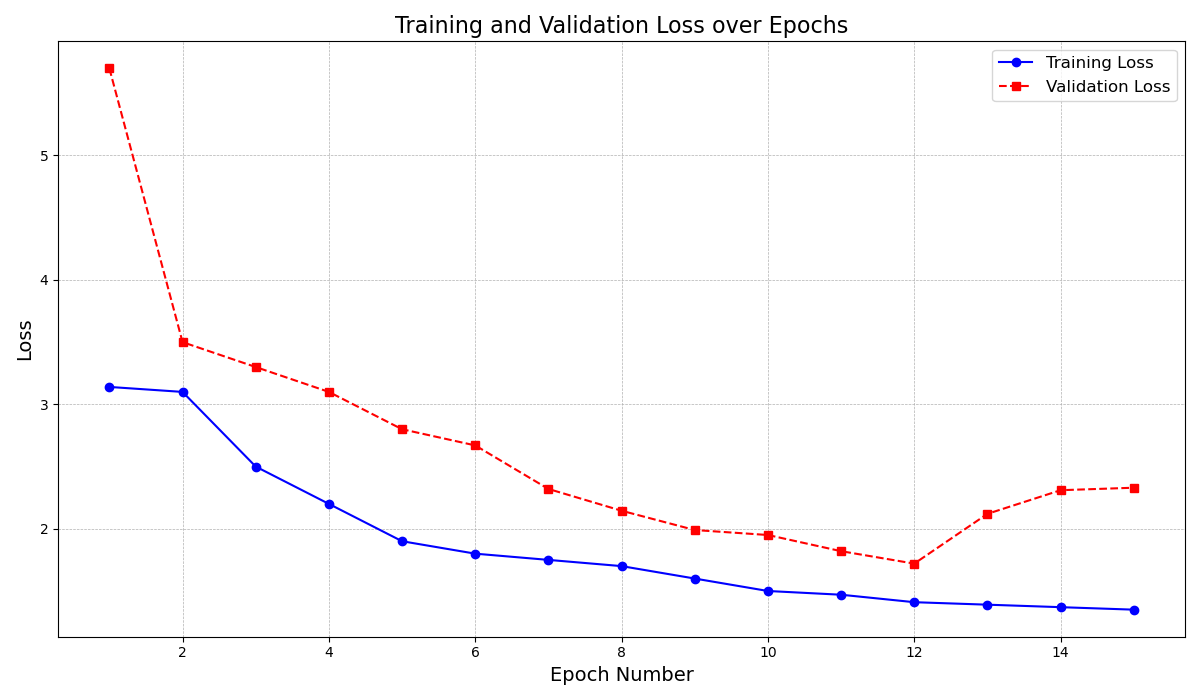}
    \caption[]{Shows training vs validation loss.}
    \label{fig:train_loss}
\end{figure}

We collected the training data as described in Section~\ref{feature} at 10Hz. During the data collection we drive the warthog for approximately 5 mins and collected 2726 samples. This data is then divided into (80,20) split of training examples and validation examples. We then performed 15 epochs of training with a learning rate of $1 \times 10^{-3}$ and batch size of 256. Fig. \ref{fig:train_loss} shows the training and validation loss with number of epochs.
We save the model checkpoints at each epoch and use the model checkpoint corresponding to the 12th epoch for the testing.
\begin{table}[h]
    \centering
    \caption{Training scenarios}
    \label{tab:sample}
    \resizebox{0.9\columnwidth}{!}{
    \begin{tabular}{|c|c|c|}
        \hline
        \textbf{Scenario} & \textbf{Terrains} & \textbf{Human preference} \\
        \hline
        1 & water(blue texture) & avoid \\
        \hline
        2 &  rock(dark grey texture) & avoid \\
        \hline
         3 & mud (dark brown texture) & avoid \\
        \hline
         4 & non traversable (bright orange texture) & avoid \\
        \hline
         5 & mud vs rock & prefer mud \\
        \hline
         6 & rock vs water  & prefer rock \\
        \hline
         7 & water vs non traversable& prefer water \\
        \hline
         8 & non traversable vs water & prefer water \\
        \hline
         9 & water vs rock & prefer rock \\
        \hline
         10 & rock vs mud & prefer mud \\
        \hline
    \end{tabular}}
\end{table}

\subsection{Testing}

\begin{figure}[htbp]
\captionsetup{font=footnotesize}
    \centering
    
    \begin{subfigure}[b]{0.23\textwidth}  
        \centering
        \includegraphics[width=\textwidth]{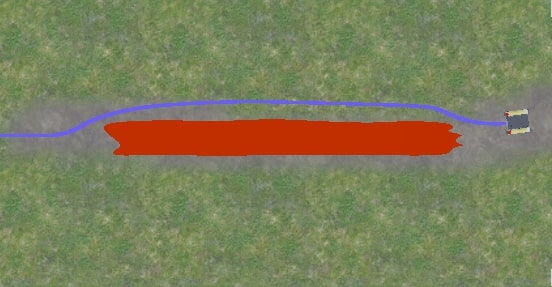}
        \caption{non traversable, longer terrain patch}
        \label{fig:sub11}
    \end{subfigure}
    \hfill
    \begin{subfigure}[b]{0.23\textwidth}
        \centering
        \includegraphics[width=\textwidth]{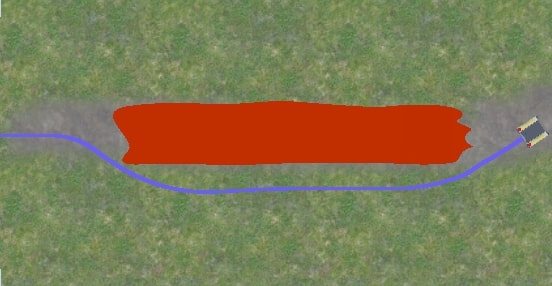}
        \caption{non traversable, wider terrin patch}
        \label{fig:sub12}
    \end{subfigure}
    \hfill
    \begin{subfigure}[b]{0.23\textwidth}
        \centering
        \includegraphics[width=\textwidth]{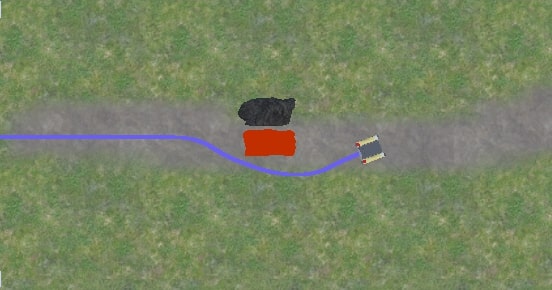}
        \caption{rock vs non traversable}
        \label{fig:sub13}
    \end{subfigure}
    \hfill
    \begin{subfigure}[b]{0.23\textwidth}
        \centering
        \includegraphics[width=\textwidth]{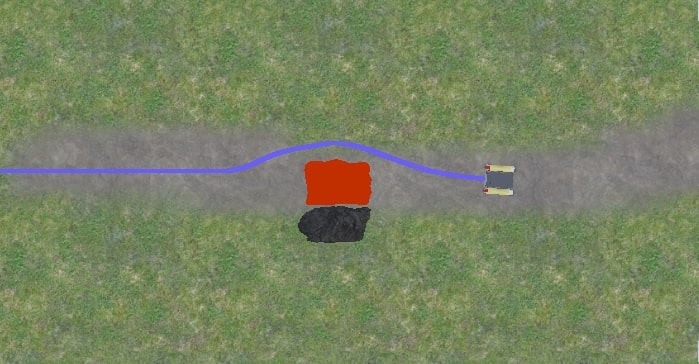}
        \caption{rock vs non traversable}
        \label{fig:sub14}
    \end{subfigure}

    \caption{Showing test results on various scenarios not present in training set}
    \label{fig:main1}
\end{figure}

\begin{figure}[htbp]
\captionsetup{font=footnotesize}
    \centering
    
    \begin{subfigure}[b]{0.23\textwidth}  
        \centering
        \includegraphics[width=\textwidth]{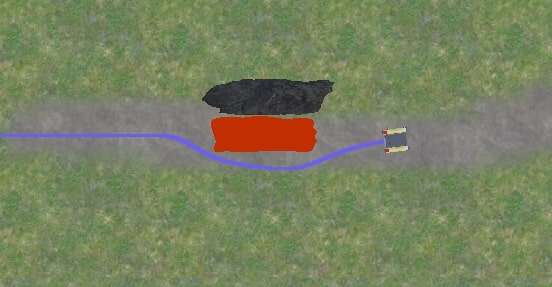}
        \caption{rock vs non traversable both wider and longer}
        \label{fig:sub21}
    \end{subfigure}
    \hfill
    \begin{subfigure}[b]{0.23\textwidth}
        \centering
        \includegraphics[width=\textwidth]{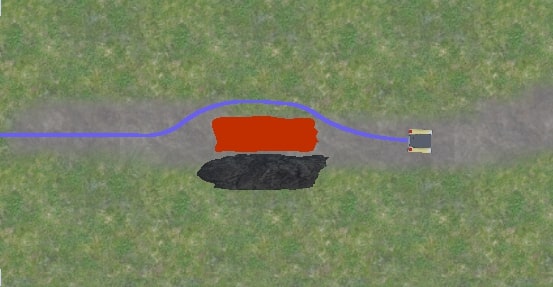}
        \caption{rock vs non traversable both wider and longer}
        \label{fig:sub22}
    \end{subfigure}
    \hfill
    \begin{subfigure}[b]{0.23\textwidth}
        \centering
        \includegraphics[width=\textwidth]{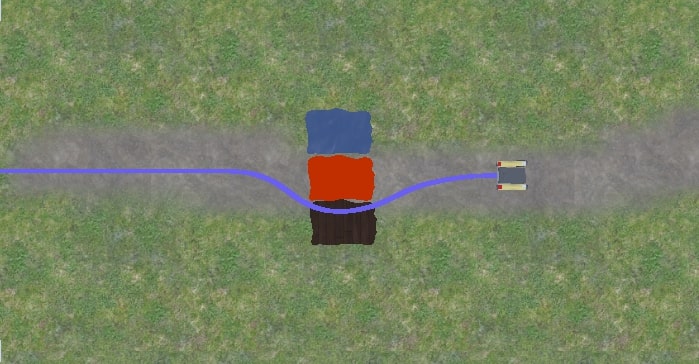}
        \caption{water vs mud vs rock, prefers mud}
        \label{fig:sub23}
    \end{subfigure}
    \hfill
    \begin{subfigure}[b]{0.23\textwidth}
        \centering
        \includegraphics[width=\textwidth]{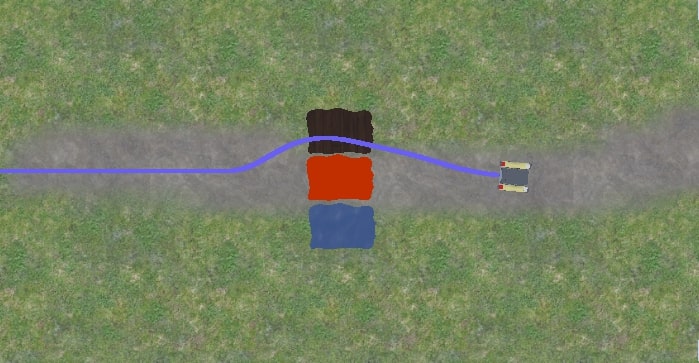}
        \caption{water vs mud vs rock, prefers mud}
        \label{fig:sub24}
    \end{subfigure}

    \caption{Showing test results on various scenarios not present in training set}
    \label{fig:main2}
\end{figure}

Fig. \ref{fig:test_sim} shows the reference path and scenarios for testing. We evaluate our planner on 7 different scenarios 
that cover a wide variety of terrain configurations in the off-road environment.
The first two scenarios have rock and water terrain types with different shape of terrain
patches than the ones used in training. The planner trajectory for this scenario is shown 
in Fig. \ref{fig:test_sim2}, as we can see from the trajectory that the warthog is able to avoid
terrain features of different sizes. The terrain shapes are also kept different from the ones
in training data of all the subsequent scenarios. Fig. \ref{fig:test_sim2} visualizes the trajectory taken by warthog for all the scenarios and Table-\ref{tab:sample2} summarizes the testing results. Fig. \ref{fig:main1} and Fig. \ref{fig:main2}, provide more results with different terrain sizes and scenarios. Fig. \ref{fig:sub11} and Fig. \ref{fig:sub12} show results on terrain with variation in length and width.  Fig. \ref{fig:sub13} and Fig. \ref{fig:sub14} shows that the planner can adapt and either go left or right
based on where the rocky terrain is. Fig. \ref{fig:sub21} and Fig. \ref{fig:sub22} repeat the same experiment but with different terrain sizes. Fig. \ref{fig:sub23} and Fig. \ref{fig:sub24} show that when presented with 3 different terrains warthog is able to choose the most preferred one (mud) and adapt (going left vs going right).
Even with training based on only five minutes of data, these results demonstrate that the planner can adapt to terrain configurations absent from the training set and accommodate variations in terrain size

\begin{table}[h]
    \centering
    
    \caption{Testing Scenarios}
    \resizebox{\columnwidth}{!}{
    
    \begin{tabular}{|c|c|c|c|}
        \hline
        \textbf{Scenario} & \textbf{Terrains} & \textbf{Human preference} & \textbf{planner preference} \\
        \hline
        1 & water & avoid & avoid\\
        \hline
        2 &  rock & avoid & avoid\\
        \hline
         3 & mud, non traversable & prefer mud & prefer mud \\
        \hline
         4 & water, non traversable, mud & prefer mud & prefer mud \\
        \hline
         5 & mud, rock, water & prefer mud & prefer mud\\
        \hline
         6 & mud, water inside & prefer mud & prefer mud\\
        \hline
         7 & rock, non traversable  & prefer rock & prefer rock \\
        \hline
    \end{tabular}}
    \label{tab:sample2}
\end{table}

\section{CONCLUSIONS}
In this work we presented a local planning algorithm that uses monocular camera to learn human driving preferences in off-road settings. We presented various experiments in simulation to show the effectiveness of the algorithm. The planner shows quick learning and adaptation, requiring only 5-10 minutes of demonstration data to effectively navigate in challenging off-road environments. Unlike current approaches that rely on extensive labeled datasets or precise sensor calibration, our method can generalize to new terrain configurations not seen during training. This ability to adapt makes our approach particularly suitable for the diverse and unpredictable nature of off-road navigation.

The success of our local planner in accurately emulating human driving preferences in off-road scenarios, coupled with its quick adaptability, makes it an attractive candidate for autonomous vehicle navigation in challenging terrains. This research paves the way for more intuitive and human-like autonomous driving solutions but also significantly reduces the barrier to entry for deploying such technologies in diverse off-road environments. Future work will focus on validating the planner's performance in real-world off-road settings to assess its sim-to-real transfer capabilities along with expanding the range of terrain types, incorporating additional sensor modalities, and integrating with higher-level path planning algorithms to enhance its applicability in real-world autonomous systems

\bibliography{ifacconf}             

\end{document}